\documentclass[11pt,a4paper]{article}
\usepackage{times,latexsym}
\usepackage{url}
\usepackage[T1]{fontenc}
\usepackage{multirow}
\usepackage{amsmath}
\usepackage{amssymb}
\usepackage{soul}
\usepackage{amsfonts}
\usepackage{booktabs}
\usepackage{siunitx}
\usepackage{tabularx}
\usepackage{appendix}
\makeatletter
\newcommand\notsotiny{\@setfontsize\notsotiny{7.3}{5}}
\makeatother
\newcolumntype{L}{>{\raggedright\arraybackslash}X}
\newcolumntype{R}{>{\raggedleft\arraybackslash}X}
\usepackage{graphicx}

\usepackage[acceptedWithA]{tacl2018v2}

\newcommand{\oneS}{\ensuremath{{}^{\textstyle *}}}

\newcommand{\threeS}{\ensuremath{{}^{\textstyle **}\oneS}}

\title{Morphology Matters: A Multilingual Language Modeling Analysis}

\author{Hyunji Hayley Park \\ University of Illinois \\  {\sf hpark129@illinois.edu }
         \And  Katherine J. Zhang \\ Carnegie Mellon University \\ {\sf kjzhang@cmu.edu } \And
         Coleman Haley \\ Johns Hopkins University \\ {\sf chaley7@jhu.edu } \AND
         Kenneth Steimel \\ Indiana University \\ {\sf ksteimel@iu.edu }\\ \And 
         Han Liu \\  University of Chicago\thanks{\textsuperscript{ } Work done while at University of Colorado Boulder}\\ {\sf hanliu@uchicago.edu } \And
         Lane Schwartz \\ University of Illinois \\ {\sf lanes@illinois.edu } }

\date{}

\begin{document}
\maketitle
\begin{abstract}

Prior studies in multilingual language modeling \cite[e.g.,][]{cotterell-etal-2018-languages, mielke-etal-2019-kind} disagree on whether or not inflectional morphology makes languages harder to model.
We attempt to resolve the disagreement and extend those studies. We compile a larger corpus of 145 Bible translations in 92 languages and a larger number of typological features.\footnote{\url{https://github.com/hayleypark/MorphologyMatters}} We fill in missing typological data for several languages
and consider corpus-based measures of morphological complexity in addition to expert-produced typological features.
We find that several morphological measures are significantly associated with higher surprisal when LSTM models are trained with BPE-segmented data. 
We also investigate linguistically-motivated subword segmentation strategies like Morfessor and Finite-State Transducers (FSTs) and find that these segmentation strategies yield better performance and reduce the impact of a language's morphology on language modeling.

\end{abstract}

\section{Introduction}

With most research in Natural Language Processing (NLP) directed at a small subset of the world's languages, whether the techniques developed are truly language-agnostic is often not known. Because the vast majority of research focuses on English with Chinese a distant second \cite{Mie2016Language}, neither of which is morphologically rich, the impact of morphology on NLP tasks for various languages is not entirely understood.

Several studies have investigated this issue in the context of language modeling by comparing a number of languages, but found conflicting results. \citet{gerz-etal-2018-relation} and \citet{cotterell-etal-2018-languages} find that morphological complexity is predictive of language modeling difficulty, while \citet{mielke-etal-2019-kind} conclude that simple statistics of a text like the number of types explain differences in modeling difficulty, rather than morphological measures.

This paper revisits this issue by increasing the number of languages considered and augmenting the kind and number of morphological features used. 
We train language models for 92 languages from a corpus of Bibles fully aligned at the verse level and measure language modeling performance using surprisal (the negative log-likelihood) per verse (see \S\ref{sec:metrics}).
We investigate how this measure is correlated with 12 linguist-generated morphological features and four corpus-based measures of morphological complexity. 

Additionally, we contend that the relation between segmentation method, morphology, and language modeling performance needs further investigation. 
Byte-Pair Encoding \cite[BPE;][]{shibata1999byte} is widely used in NLP tasks including machine translation \cite{sennrich-etal-2016-neural} as an unsupervised information-theoretic method for segmenting text data into subword units.
Variants of BPE or closely related methods such as WordPiece \citep{kudo-2018-subword} are frequently employed by state-of-the-art pretrained language models \citep{roberta, radford2019language, bert, xlnet}. 
However, BPE and other segmentation methods may vary in how closely they capture morphological segments for a given language, which may affect language modeling performance. 

Therefore, this paper focuses on the following two research questions:
\begin{enumerate}
\item Does a language's morphology influence language modeling difficulty?
\item If so, how do different segmentation methods interact with morphology?
\end{enumerate}

In order to answer the first question, we train models using data sets segmented by characters and BPE units.
Our results show that BPE language modeling surprisal is significantly correlated with measures of morphological typology and complexity. This suggests that BPE segments are ineffective in mitigating the effect of morphology in language modeling. 

As for the second question, we consider more linguistically-motivated segmentation methods to compare with BPE: Morfessor \cite{morfessorcreutzUnsupervisedModelsMorpheme2007} and Finite-State Transducers (FSTs) (see \S\ref{sec:segmentation}).
Our comparison of the models using the different segmentation methods shows that Morfessor reduces the impact of morphology for more languages than BPE. FST-based segmentation methods outperform the other segmentation methods when available. 
These results suggest that morphologically motivated segmentations improve cross-linguistic language modeling.

\section{Modeling difficulty across languages}
\label{sec:lm-review}

Studies have demonstrated that different languages may be unequally difficult to model and have tested the relations between such modeling difficulty and morphological properties of languages, using different segmentation methods.

\citet{VaniaandLopez} compared the effectiveness of word representations based on different segmentation methods in modeling 10 languages with various morphological typologies. 
They trained word-level language models, but utilize segmentation methods to create word embeddings that include segment-level information.
Comparing character, BPE, and Morfessor segmentations, they concluded that character-based representations were most effective across languages, with BPE always outperforming Morfessor. However, models based on hand-crafted morphological analyses outperformed all other segmentation methods by a wide margin.

\citet{gerz-etal-2018-relation} trained n-gram and neural language models over 50 languages and argued that the type of morphological system is predictive of model performance.
Their results show that languages differ with regard to modeling difficulty.
They attributed the differences among languages to four types of morphological systems: isolating, fusional, introflexive, and agglutinative. 
While they found a significant association between the morphological type and modeling difficulty, Type-Token Ratio (TTR) was the most predictive of language modeling performance. 

\citet{cotterell-etal-2018-languages} arrived at a similar conclusion modeling 21 languages using the Europarl corpus \citep{koehn2005europarl}.
When trained with n-gram and character-based Long Short-Term Memory (LSTM) models, the languages showed different modeling difficulties, which were correlated with a measure of morphology, Morphological Counting Complexity (MCC) or the number of inflectional categories \citep{sagot-complexity}. 

However, \citet{mielke-etal-2019-kind} failed to reproduce the correlation with MCC when they increased the scope to 69 languages, utilizing a Bible corpus \citep{mayer-cysouw-2014-creating}.
They also reported no correlation with measures of morphosyntactic complexity such as head-POS entropy \citep{dehouck-denis-2018-framework} and other linguist-generated features \citep{wals}. 
Rather, they found that simpler statistics, namely the number of types and number of characters per word, correlate with language model surprisal using BPE and character segmentation, respectively.

\section{Morphological measures}

Different measures of morphology are used to represent a language's morphology.

\subsection{Linguist-generated measures}
The most linguistically-informed measures of morphology involve expert descriptions of languages. 
The World Atlas of Language Structures \citep[WALS;][]{wals} has been used frequently in the literature to provide typological information. 
WALS is a large database of linguistic features gathered from descriptive materials, such as reference grammars. It contains 144 chapters in 11 areas including phonology, morphology, and word order. Each chapter describes a feature with categorical values and lists languages that have each value.
However, not all languages in the database have data for all the features, and for some languages there is no data at all. 

The studies reviewed in \S\ref{sec:lm-review} all relied on this expert-description approach to quantify morphological properties.
\citet{gerz-etal-2018-relation} focused on WALS descriptions of inflectional synthesis of verbs, fusion, exponence, and flexivity, while \citet{mielke-etal-2019-kind} looked at two WALS features, 26A ``Prefixing vs. Suffixing in Inflectional Morphology'' and 81A ``Order of Subject, Object and Verb.''
\citet{cotterell-etal-2018-languages} used UniMorph \citep{kirov-etal-2018-unimorph}, instead of WALS, to calculate MCC.
\citet{VaniaandLopez} did not cite any databases but provided descriptions of four morphological types (fusional, agglutinative, root-and-pattern, and reduplication) and categorized 10 languages into these types.

A major issue with this approach to representing morphology is that there is not enough expert data available to enable comparisons across many different languages. 
In fact, \citet{mielke-etal-2019-kind} chose their two WALS features because data for these features existed for most of their languages. 
Moreover, \citet{bentzComparisonMorphologicalComplexity2016c} showed that their WALS-based measure had lower correlations with other measures of morphological complexity due to this issue of missing data.

\subsection{Corpus-based measures}
In contrast, corpus-based measures of morphology can be easily calculated on a given data set.
These measures include the number of types, Type-Token Ratio (TTR), Moving-Average TTR \citep[MATTR;][]{covingtonCuttingGordianKnot2010}, and Mean Length of Words (MLW).
The exact definition of the measures may vary depending on studies, but we define them as in Table \ref{table1:corpus-based-measures}, 
where a word token is a string separated by spaces in the training set after tokenization but before segmentation.

While some studies \citep[e.g.,][]{mielke-etal-2019-kind} consider these measures as simple statistics of a corpus, other studies have found that they can be used as approximate measures of morphological complexity. \citet{kettunenCanTypeTokenRatio2014a} showed that TTR, MATTR, and MLW can capture the overall ranking of morphological complexity generated by information-theoretic and expert-generated measures of morphological complexity.
\citet{bentzComparisonMorphologicalComplexity2016c} compared different measures of morphological complexity for 519 languages across 101 families and showed a strong correlation between all measures, which were based on corpus statistics, linguistic expertise, information theory, and translation alignment.
They argued that corpus-based measures, including TTR, and other measures of morphological complexity can be used interchangeably.
In addition, \citet{gerz-etal-2018-relation} showed that TTR is influenced by the morphological typology of a language. 
According to them, isolating languages tend to have small TTR values and are often easier to model while the opposite is true for agglutinative languages.

\begin{table}[t!]
\centering
 \begin{tabularx}{\columnwidth}{lX}
\toprule
Measure & Definition \\
\midrule
Types & Number of unique word tokens\\
TTR & Number of unique word tokens divided by total number of word tokens\\
MATTR & Average TTR calculated over a moving window of 500 word tokens\\
MLW & Average number of characters per word token\\
\bottomrule
\end{tabularx}
\caption{Corpus-based measures of morphology defined for this study. These measures are calculated on tokenized data sets before applying any segmentation method.}
\label{table1:corpus-based-measures}
\end{table}

Given the previous literature, we utilize these corpus-based measures, as well as expert-generated WALS features, as a proxy for morphological differences among languages in our study. 

\section{Methods}

We design our experiments to test if a language's morphology is correlated with language model performance, depending on the segmentation method. 
We represent a language's morphology using WALS features and corpus statistics. 
We train language models for Bible translations in 92 languages based on five different segmentation methods: character, BPE, Morfessor, and FST with BPE or Morfessor back-off strategies (FST$+$BPE \& FST$+$Morfessor). 
We use surprisal per verse \cite{mielke-etal-2019-kind} as the evaluation metric to compare language modeling performance across different languages and different segmentation methods. Additionally, we quantify the difference in surprisal per verse between segmentation methods to compare the relative strength of each segmentation method with regard to morphological complexity.

\subsection{Data}

Our data consist of 145 Bible translations in 92 languages covering 22 language families,\footnote{For each language, we report the family assigned by WALS \citep{wals}: 6 Afro-Asiatic, 1 Algic, 1 Altaic, 2 Austro-Asiatic, 6 Austronesian, 1 Aymaran, 3 Dravidian, 4 Eskimo-Aleut, 1 Guaicuruan, 33 Indo-European, 1 Japanese, 1 Korean, 1 Mande, 6 Mayan, 6 Niger-Congo, 4 Quechuan, 5 Sino-Tibetan, 1 Songhay, 1 Tai-Kadai, 2 Tupian, 2 Uralic, 2 Uto-Aztecan, 2 creoles.} fully aligned at the verse level.
The majority of the data came verse-aligned from \citet{mielke-etal-2019-kind} \citep[original data from][]{mayer-cysouw-2014-creating}. We added more Bibles from another corpus \citep{bible-Christo-2014} and from online Bible resources (see Appendix \ref{Appendix:data} for more information).
We refer to each language by ISO 639-3 code when applicable.

We followed \citet{mielke-etal-2019-kind}'s method to split the data into training, development, and test sets:
the verse-aligned data were divided into blocks of 30 verses, with the first five verses being assigned to the development set, the next five to the test set and the rest to the training set.
The resulting training set had 16,926 verses while development and test sets had 4,225 verses each.

It should be noted that both \citet{mielke-etal-2019-kind} and \citet{bible-Christo-2014} provided tokenized data. We tokenized the newly added Bibles using \citet{mielke-etal-2019-tokenizer}'s tokenizer, following \citet{mielke-etal-2019-kind}. 
When both tokenized and untokenized versions were available, we included the tokenized versions only.

We chose to replace characters that only occurred one time with a special \texttt{UNK} symbol.
\citet{mielke-etal-2019-kind} applied this procedure to characters that appear less than 25 times in the training set except for Chinese, where only  singleton characters were replaced. 
Because we added several languages where the original strategy would have resulted in removing too much data, we preprocessed singleton characters across the board.

We also corrected several errors present in the data.
For example, the Bible translations in Shona ({sna}) and Telugu ({tel}) were mis-coded as Shan ({shn}) and Tecpatlán Totonac ({tcw}), respectively.

\subsection{Morphological measures selected}
In this paper, we adopt two approaches to representing a language's morphology. First, we rely on expert descriptions of languages in WALS, manually augmenting the database to rectify the issue of missing data.
Second, we utilize corpus-based measures like TTR to represent the morphological complexity of a given language.

\paragraph{WALS features}

While some previous studies \citep[e.g.,][]{gerz-etal-2018-relation, VaniaandLopez} categorized relatively well-known languages into a small number of morphological types, such categorization is not always clear. 
Some other studies \citep[e.g.,][]{cotterell-etal-2018-languages, mielke-etal-2019-kind} selected a small number of available typological features to compare, but their conclusions were at odds, possibly calling for exploration of other measures.
Therefore, we consider all available morphological features described by WALS to explore which features affect language modeling and how.
Instead of making theoretical claims about morphological typology, we explore which typological features make a language's morphology more complex for LSTM language models. 
 
To that end, we augmented the existing WALS database by consulting reference grammars for each language.
Of the 92 languages in our corpus, six were not in the WALS database.\footnote{{ikt, lat, nch, tbz, wbm, zom}} In addition, many of the languages in the database had missing data for some features. For example, we had no data for any of the morphological features of Afrikaans ({afr}). We manually assigned missing features where possible following the descriptions in the relevant WALS chapters regarding the procedures used to assign feature values to languages.

\begin{table}[t!]
\centering
 \begin{tabularx}{\columnwidth}{lX}
\toprule
ID & Name \\
\midrule
20A & Fusion of Selected Inflectional Formatives \\
21A & Exponence of Selected Inflectional Formatives \\
21B & Exponence of Tense-Aspect-Mood Inflection \\
22A & Inflectional Synthesis of the Verb \\
23A & Locus of Marking in the Clause \\
24A & Locus of Marking in Possessive Noun Phrases \\
25A & Locus of Marking: Whole-language Typology \\
25B & Zero Marking of A and P Arguments \\
26A & Prefixing vs. Suffixing in Inflectional Morphology \\
27A & Reduplication \\
28A & Case Syncretism \\
29A & Syncretism in Verbal Person/Number Marking \\
\bottomrule
\end{tabularx}
\caption{The 12 morphological features in WALS.}
\label{table2:wals-features}
\end{table}

Of the almost 200 features in WALS, the editors of the database labeled 12 of them as morphological features. Therefore, we considered these 12 features, listed in Table \ref{table2:wals-features} and described below,\footnote{See \url{https://wals.info/chapter} for more details and examples of these features.} to test the hypothesis that morphological complexity correlates with modeling difficulty.

Feature 20A describes how closely grammatical markers (inflectional formatives) are phonologically connected to a host word or stem. The markers can be isolating, concatenative, or even nonlinear (i.e., ablaut and tone).

Features 21A and 21B measure the exponence of selected grammatical markers. Exponence refers to the number of categories that a single morpheme expresses. For 21A, the selected grammatical markers were case markers. For 21B, they were tense-aspect-mood (TAM) markers.

Feature 22A measures how many grammatical categories may appear on verbs in a language. These categories include tense-aspect-mood, negation, voice, and agreement.

Features 23A through 25B describe the existence and locus of marking in different kinds of phrases. A phrase may have marking on either its head, its dependent(s), both, or neither. In full clauses, the verb is the head, and the subject and object arguments are dependents. In possessive noun phrases, the possessed noun is the head while the possessor is dependent.

Feature 26A measures the degree to which languages use prefixes versus suffixes in their inflectional morphology. Feature 27A describes which languages use reduplication productively and whether or not both full and partial reduplication are used.

Both Features 28A and 29A measure syncretism. Syncretism occurs when a single inflected form corresponds to more than one function. 28A measures case syncretism specifically while 29A measures syncretism in the subject agreement marking of verbs.

\paragraph{Types, TTR, MATTR, and MLW}
We calculated the number of types, TTR, Moving-Average TTR, and Mean Length of Word using an adapted script from the Python module LexicalRichness.\footnote{\url{https://github.com/LSYS/LexicalRichness}}
We used a window size of 500 for Moving-Average TTR, following previous studies \citep[e.g.,][]{kettunenCanTypeTokenRatio2014a}.
The definitions of the measures are found in Table \ref{table1:corpus-based-measures}. All measures were calculated based on the word tokens in the training set before applying any segmentation method.

\begin{table*}[t]
\hspace*{-4pt}
\centering
 \begin{tabularx}{1.003\textwidth}{l|X}
\toprule
Segmentation & Example \\
\midrule
Tokenized & {\small \texttt{Yuhannan{\i}n karde\c{s}i Yakubu k{\i}l{\i}\c{c}la \"{o}ld\"{u}rd\"{u} .}} \\
Character & {\notsotiny \texttt{Y u h a n n a n {\i} n \_ k a r d e \c{s} i \_ Y a k u b u \_ k {\i} l {\i} \c{c} l a \_ \"{o} l d \"{u} r d \"{u} . }}\\
BPE & {\small \texttt{Yuhan@@ nan{\i}n karde\c{s}i Yakubu k{\i}l{\i}\c{c}la \"{o}ld\"{u}rd\"{u} . }}  \\
Morfessor & {\small \texttt{Yuhanna@@ n{\i}n karde\c{s}@@ i Yakub@@ u k{\i}l{\i}\c{c}@@ la \"{o}ld\"{u}rd\"{u} . }} \\
\small{FST$+$BPE} & {\small \texttt{Yuhan@@ nan{\i}n karde\c{s}@@ i Yakub@@ u k{\i}l{\i}\c{c}@@ la \"{o}l@@ d\"{u}r@@ d\"{u} . }}\\
\small{FST$+$Morfessor} & {\small \texttt{Yuhanna@@ n{\i}n karde\c{s}@@ i Yakub@@ u k{\i}l{\i}\c{c}@@ la \"{o}l@@ d\"{u}r@@ d\"{u} . }}\\
\bottomrule
\end{tabularx}
\caption{Turkish examples for different segmentation methods. An English translation is ``And he killed James the brother of John with the sword'' (Acts 12:2). FST does not produce analyses for \textit{Yuhannan{\i}n} (``John's''), for which BPE or Morfessor back-off was used. The segmentation created by human experts was the same as FST$+$Morfessor. $\langle@@\rangle$ denotes subword segmentation while $\langle\_\rangle$ encodes space between word tokens for character segmentation.}
\label{table3:segmentation-sample}
\end{table*}

\subsection{Segmentation methods}\label{sec:segmentation}
We chose to train only open-vocabulary language models for fair comparison.
Word-level models will predict \texttt{UNK} for out-of-vocabulary word tokens and cannot be fairly compared with character- and subword-level models as a result. 
Specifically, we trained language models using five segmentation methods: character, BPE, Morfessor, FST$+$BPE, and FST$+$Morfessor. These segmentation methods provide a way to segment any given text into smaller pieces, some of which approximate morphemes.

A morpheme is the smallest meaning-bearing morphological unit while a morph is the surface representation of one or more morphemes. Linguistically-motivated methods like Morfessor and FSTs are designed with the goal of producing subword segments that are closely aligned to the true morphs comprising a word. While BPE was not designed with morpheme segmentation in mind, its resulting subwords are commonly believed to align with morphs to some degree due to morph subsequences being frequent in the data.

Segmenting words into morphs may reduce the impact of rich morphology as highly inflected words can be broken into smaller pieces that are likely to contribute similar meanings across contexts in the corpus.
Table \ref{table3:segmentation-sample} provides examples of the segmentation methods we used to train language models.
The original verse is provided for reference only and not used to train any models.

\paragraph{Character}
We trained character-based language models, following previous studies \citep{mielke-etal-2019-kind, gerz-etal-2018-relation, cotterell-etal-2018-languages}.
Character language models are trained to predict the next character given the preceding context, and the vocabulary includes an underscore $\langle\_\rangle$ to denote word boundaries.

\paragraph{BPE} We trained BPE-based language models, following \citet{mielke-etal-2019-kind}.
Starting with character segmentation, BPE operations combine characters into larger chunks based on their frequencies to create units somewhere between characters and words with the number of merge operations as the hyperparameter \citep{sennrich-etal-2016-neural}.
We used $0.4\times\mathit{types}$
as the number of merges, as \citet{mielke-etal-2019-kind} reported that to be most effective with their corpus.\footnote{Additional static numbers of merge operations were also tested, with nearly identical results.}
BPE language models are trained to predict the next BPE unit. The double at sign $\langle@@\rangle$ is used to indicate segments that are not word-final.

\paragraph{Morfessor} 
Morfessor \cite{morfessorcreutzUnsupervisedModelsMorpheme2007} is a word segmentation method explicitly designed for morphological segmentation.
The default implementation utilizes a unigram language model to find morph-like constructs.
While like BPE this approach is information-theoretic, it selects segments top-down and includes a prior term for the length of segments, regularizing segments to be more plausible morphemes.

Using the default settings with Morfessor 2.0 \cite{Virpioja2013Morfessor2P}, we trained Morfessor on the training set and applied the segmentation to all data sets. 
Just like BPE, the language models are trained to predict the next morph unit.

\paragraph{FST} \label{sec:fst_seg} While segmentation based on BPE and Morfessor may or may not resemble actual morphemes, morpheme segmentation from Finite-State Transducers (FSTs) provides a knowledge-based method to segment a text into morphemes.
Finite-state morphological analyzers are rule-based systems that take a surface string as input and produce all possible morphological analyses as output.
To use FSTs for segmentation, we changed existing morphological analyzers into segmenters and developed a heuristic to select one analysis for a given word token.
FSTs for Plains Cree \cite{arppe2019finite}, German \cite{schmid2004smor}, English \cite{eng_helsinki}, Finnish \cite{pirinen2015omorfi}, Indonesian \cite{larasati2011indonesian}, Cuzco Quechua \cite{quz-fst}, and Turkish \cite{Coltekin2014, Coltekin2010} were used as morphological segmenters.

Most FSTs are designed to provide analyses for surface forms, not morphological segmentations.
Fortunately, morpheme boundaries are frequently part of FSTs due to their relevance for lexico-phonological phenomena. 
By modifying the FST before the cleanup rules that remove morpheme boundaries can apply, we create a morphological segmenter that takes in a surface form and returns the surface form with morpheme boundary markers.
If the analyzer provides segmentations, the transducer is used as-is.

For example, the Turkish FST produces a morphological analysis for the surface form \textit{k{\i}l{\i}\c{c}la} (``with the sword") in the example in Table \ref{table3:segmentation-sample}: \texttt{k{\i}l{\i}\c{c}<NOUN><Case:instrumental>}. Instead of producing such an analysis for the given word, the segmenter instead produces the segmented surface form \texttt{k{\i}l{\i}\c{c}@@ la}, which is used in the FST segmentation methods.  

Because a FST may return multiple analyses or segmentations given a single word, a heuristic method was used to determine which segmentation to select. In general, we chose the segmentation with the fewest segments. However, the English segmenter based on \citet{eng_helsinki} always returns the input string itself as a possible segmentation if covered by the analyzer. For example, \textit{walks} would produce two segmentations in the English segmenter: \texttt{walks} and \texttt{walk@@ s}. For this segmenter, we selected the fewest number of segments excluding the input string itself (e.g., choosing \texttt{walk@@ s} over \texttt{walks}). 

When a FST produces no analyses for a given word, as in the case of \textit{Yuhannan{\i}n} (John's) in Table \ref{table3:segmentation-sample}, we adopt the FST-augmented BPE segmentation (FST$+$BPE) and FST-augmented Morfessor segmentation (FST$+$Morfessor), where we fall back to BPE or Morfessor segmentation whenever FST segmentation is unavailable. 
As shown in the table, FST$+$BPE and FST$+$Morfessor only differ in the segmentation of the unanalyzed word. 
For this particular verse, the human segmentation agrees with the FST$+$Morfessor segmentation. 
FST$+$BPE and FST$+$Morfessor models are trained just like BPE or Morfessor models to predict the next subword unit.

\subsection{Models} 

Following \citet{mielke-etal-2019-kind}, we trained Long Short-Term Memory (LSTM) models introduced by \citet{merityAnalysis} for each of the segmentation methods.
Three LSTM models using character, BPE, and Morfessor segmentation were trained for all languages. 
For a select group of languages, we also trained models using FST+BPE and FST+Morfessor units.
The neural architecture consisted of an initial embedding layer, multiple LSTM layers, and a linear decoder layer. For our particular experiments, we adopted the hyperparameters from \citet{mielke-etal-2019-kind} \citep[see][for their character PTB setttings]{merityAnalysis}.
The batch size used for character models was 128 with 500 epochs of training. All other models used a batch size of 40 and were trained for 200 epochs.

\subsection{Metrics} \label{sec:metrics}

\paragraph{Surprisal per verse} One major evaluation metric for language models is the negative log-likelihood on a test set. The negative log-likelihood, or surprisal, is the amount of information a language model needs to generate the next unit. 
Following \citet{mielke-etal-2019-kind}, we define the surprisal at the verse level, where $\mathrm{NLL}(v_{ij}) = -\log_2 p(v_{ij})$ with a verse $v_{ij}$ (for $i$th verse in language $j$).
Since each verse is intended to express the same meaning across languages, differences in per-verse surprisal across languages primarily indicate differences in cross-linguistic language model quality (rather than differences in meaning content).

For each language $j$, we average the negative log-likelihood across the 4,225 verses in the test set, making 
$\mathrm{L}_j = \frac{1}{4225} \sum_{i=1}^{4225} \mathrm{NLL}(v_{ij})$.

\paragraph{Surprisal difference} Additionally, we quantify the difference between segmentation methods in language modeling performance as shown in Equation \ref{eq:delta}. This quantity compares the relative strength of one segmentation method to another.
\begin{equation}
\mathrm{}{\Delta_{S_{j1}, S_{j2}}} = \frac{L_{j1} - L_{j2}}{\frac{1}{2}(L_{j1} + L_{j2})}
\label{eq:delta}
\end{equation}
$S_{j1}$ and $S_{j2}$ are two segmentation methods to compare and
$L_{j1}$ and $L_{j2}$ represent the surprisal per verse for the language models based on the two segmentation methods. 
If $\Delta_{S_{j1}, S_{j2}}$ is positive, $S_{j1}$ resulted in a higher surprisal than $S_{j2}$ and $S_{j2}$ was more effective in modeling a given language.

\section{Results}
\label{sec:results}
We now present results from our experiments. We report the strong association between several morphological features and surprisal per verse for BPE language models, compared to language models based on other segmentation methods. Then, we show the trade offs between different segmentation methods and how they interact with morphological complexity. 
Our assumption is that, if a segmentation method reduces the impact of morphology, the surprisal values of language models based on that segmentation will have weaker correlations with measures of morphology.

\subsection{Correlation studies with character and BPE models}

We investigated correlations between surprisal per verse and various measures of morphology (i.e., WALS features, number of types, TTR, Moving-Average TTR, Mean Length of Word).
\citet{benjamini1995controlling}'s procedure was used to control the false discovery rate, so only $p\leq\frac{8}{15}\cdot0.05$ ($\approx0.027$) is considered significant.

\paragraph{WALS features} 
We tested for association between surprisal and each selected WALS feature with the Kruskal--Wallis test, or one-way ANOVA on ranks. This non-parametric test was chosen because the distribution of surprisal values did not meet the assumption of normality.
A significant test result in this context means that there are significant differences in the median surprisal values between categories for a given feature.
In order for the test to be effective, only feature values with a sample size $\geq5$ were tested.

For the character models, no features showed significant association with surprisal. However, for the BPE models, half of the morphological features had significant association with surprisal. 
These features were 21A ``Exponence of Selected Inflectional Formatives,'' 23A ``Locus of Marking in the Clause,'' 24A ``Locus of Marking in Possessive Noun Phrases,'' 25A ``Locus of Marking: Whole-language Typology,'' 25B ``Zero Marking of A and P Arguments,'' and 29A ``Syncretism in Verbal Person/Number Marking.''

For the features shown to have an effect on the BPE surprisal, we calculated the effect sizes and performed post-hoc comparisons to determine which categories were significantly different.
In this context, effect size ($\eta^2$) indicates the proportion of variance in surprisal per verse explained by each WALS feature, and $\eta^2 \geq 0.14$ is considered a large effect \cite{tomczakNeedReportEffect2014}.
The $p$-values and effect sizes are summarized in Table \ref{table4:wals-effect-size}.
The effect size was large for all of the significant features except for 25B.

\begin{table}[t]
\centering
 \begin{tabularx}{\columnwidth}{lXRS[table-format=1.3]}
\toprule
Segmentation & ID & {$p$-value} &{$\eta^2$}\\
\midrule
\multirow{4}{*}{BPE} & 21A & 1.3e-05 & \bfseries 0.28\\
& 23A& 6.7e-06 & \bfseries 0.28 \\
& 24A& 2.2e-04 & \bfseries 0.228\\
& 25A& 	6.5e-05 & \bfseries 0.253\\
& 25B& 0.014 & 0.06\\
& 29A& 2.0e-04 & \bfseries 0.198\\
\midrule
\multirow{4}{*}{Morfessor} & 21A& 0.009 & 0.109\\
& 23A& 0.002 & 0.135\\
& 26A& 0.022 & 0.064\\
& 29A& 0.024 & 0.072\\
\bottomrule
\end{tabularx}
\caption{$p$-values and effect sizes of WALS features that showed significant effect on surprisal per verse. Large effect sizes ($\geq0.14$) are in bold.}
\label{table4:wals-effect-size}
\end{table}

For Feature 21A, the median surprisal value for languages with no case was significantly lower than the median value for other types. 
Similarly, for 23A, the median surprisal value for languages with no marking was significantly lower than the value for other types. In the cases of both 24A and 25A, languages with double marking had higher surprisal values than those with single or no marking. For 25B, languages with non-zero marking had slightly higher surprisal values than those with zero-marking.
Lastly, for 29A, languages without syncretism had higher surprisal values than those with syncretism or with no marking.

In general, less inflectional morphology was associated with lower surprisal while more inflectional morphology was associated with higher surprisal.

\paragraph{Corpus-based measures}

A similar trend emerged for corpus-based measures of morphological complexity. The surprisal per verse of BPE models was highly correlated with type count, TTR, Moving-Average TTR (MATTR), and Mean Length of Word (MLW).
Yet with character models, the strength of the correlation was weak and often insignificant.
These results suggest that BPE segmentation was ineffective in reducing the impact of morphological complexity.

\begin{table}[t]
\centering
 \begin{tabularx}{\columnwidth}{XXS}
\toprule
Segmentation & Measure & {Spearman's $\rho$}\\
\midrule
\multirow{4}{*}{Character} & Types & 0.19\oneS  \\
& TTR  & 0.15  \\
& MATTR  & 0.17\oneS  \\
& MLW  & 0.06  \\
\midrule
\multirow{4}{*}{BPE} & Types & 0.80\threeS\\
& TTR  & 0.76\threeS\\
& MATTR & 0.68\threeS\\
& MLW & 0.61\threeS \\
\midrule
\multirow{4}{*}{Morfessor} & Types & 0.50\threeS\\
& TTR & 0.44\threeS\\  
& MATTR & 0.39\threeS\\
& MLW & 0.30\threeS \\
\bottomrule
\end{tabularx}
\caption{Correlation between surprisal per verse per segmentation method and morphological complexity measures. \oneS $p<0.027$, \threeS $p<0.0005$.}
\label{table5:surprisal-correlation}
\end{table}

Table \ref{table5:surprisal-correlation} summarizes the correlation coefficients and corresponding $p$-values. 
For the character-based models, only the number of types and MATTR showed a significant correlation in Spearman's rank-order correlation, and those correlations were rather weak.
In contrast, the BPE models presented strong correlations with all of the corpus-based measures at any reasonable alpha value ($p< 10^{-16}$). The number of types showed the strongest correlation, followed by TTR, MATTR, and MLW in that order. 

\begin{figure*}[t!]
    \centering
    \includegraphics[width=\textwidth]{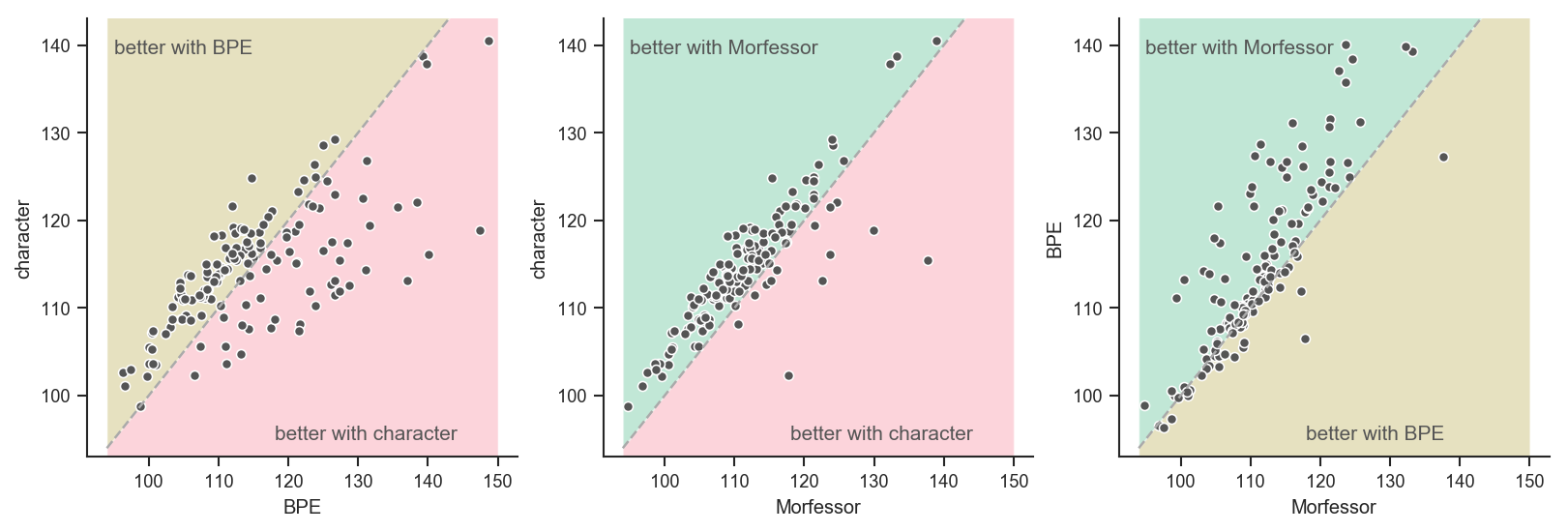}
    \caption{Pairwise comparisons of surprisal per verse values for character, BPE, and Morfessor models. For the majority of the languages, Morfessor segmentation resulted in lower surprisal per verse than character or BPE segmentation.}
    \label{figure1:pairwise_comparisons}
\end{figure*}

\subsection{Comparison with Morfessor and Finite-State Transducer models}
\label{sec:Morfessor_segmentation}

We trained language models using three additional segmentation methods: Morfessor, FST$+$BPE, and FST$+$Morfessor.
Because Morfessor is an unsupervised method, we were able to utilize it to segment all languages, but we were able to generate FST segmentation for only a few languages.
As such, we compare the character, BPE, and Morfessor models for all languages before looking into a subset of them where the FST methods were available.

\paragraph{Morfessor models}

Morfessor segmentation performed better than both character and BPE segmentation for the majority of languages. 
Figure \ref{figure1:pairwise_comparisons} shows the pairwise comparisons of the surprisal per verse values of a given language on different segmentation strategies. 
As shown in the plot on the left, the relative strength between BPE and character segmentation methods is not clear. 
BPE segmentation produced slightly better results for 49 of the 92 languages, but character segmentation produced much lower surprisal values for the rest of the languages.
In contrast, Morfessor clearly outperformed character and BPE for most of the languages, as shown in the plots in the middle and right.
Only 12 out of the 92 languages had higher surprisal values for Morfessor segmentation than character, while a total of 66 languages performed better with Morfessor segmentation than with BPE.

In addition, Morfessor models' surprisal per verse showed weaker correlations with measures of morphology.
Only four WALS features showed significant association with the Morfessor models: 21A ``Exponence of Selected Inflectional Formatives,'' 23A ``Locus of Marking in the Clause,'' 26A ``Prefixing vs. Suffixing in Inflectional Morphology,'' and 29A ``Syncretism in Verbal Person/Number Marking.''
The effect sizes were also much smaller than those for the BPE models as shown in Table \ref{table4:wals-effect-size}.

Just as with the BPE models, the median surprisal for languages with no marking was much lower than the surprisal for other types for Features 21A, 23A, and 29A.
For 26A, there was only a significant difference between weakly suffixing languages and strongly prefixing languages, with strongly prefixing languages having a lower median surprisal per verse.

As shown in Table \ref{table5:surprisal-correlation}, corpus-based statistics still showed significant correlations with the surprisal per verse value of Morfessor models, but the correlations were moderate compared to those of the BPE models.

\begin{figure}[t!]
    \centering
    \includegraphics[width=\columnwidth]{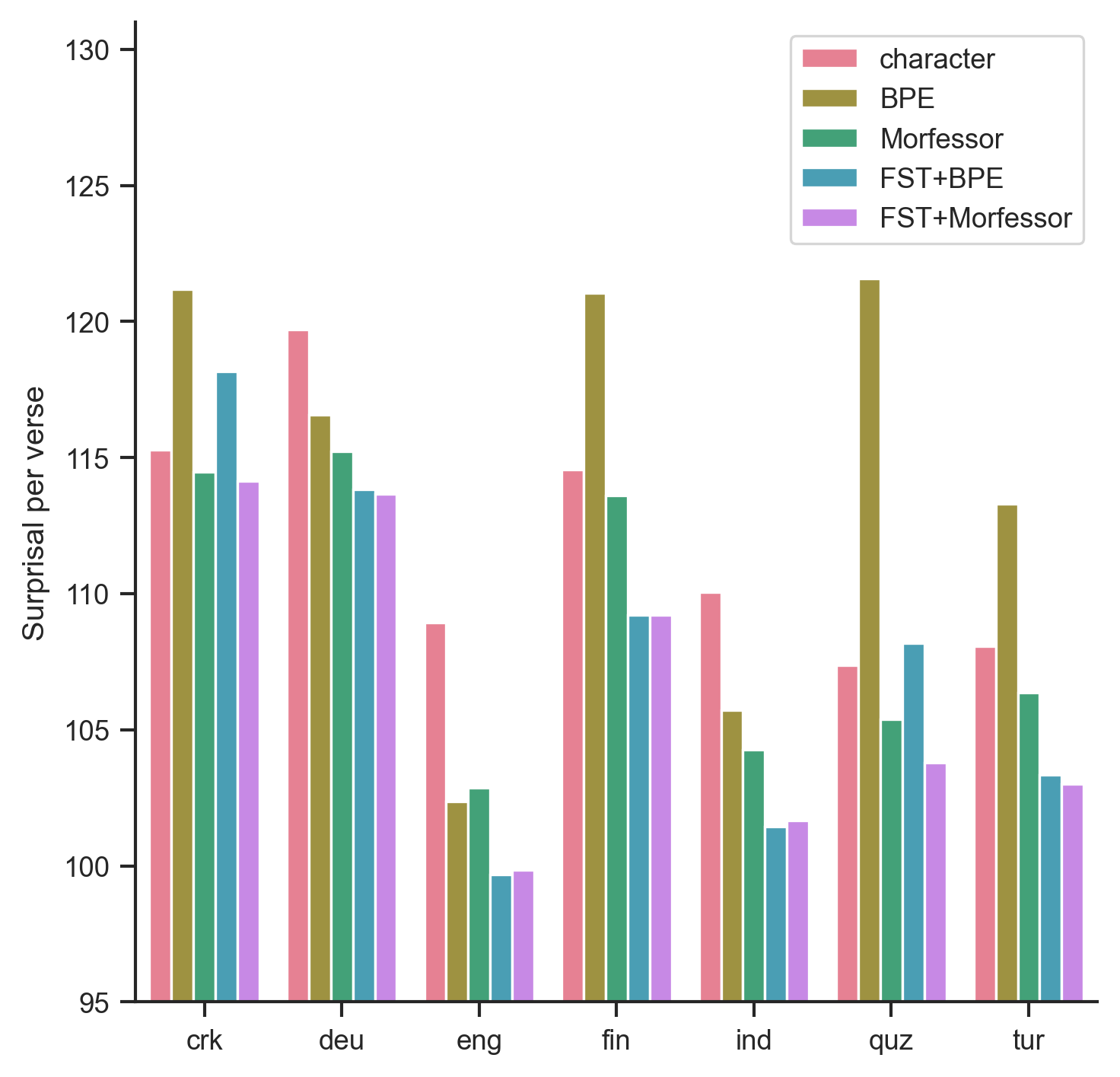}
    \caption{
    Surprisal per verse per segmentation method including FST segmentation methods. FST$+$BPE or FST$+$Morfessor models outperform all other models.
    }
    \label{figure2:fst_results}
\end{figure}

\paragraph{FST models}
When available, a FST segmentation method resulted in the best performance. The graph in Figure \ref{figure2:fst_results} displays the surprisal of FST$+$BPE and FST$+$Morfessor models in comparison to the segmentation methods discussed above.
For all seven languages, either FST$+$BPE or FST$+$Morfessor segmentation (or both) shows a clear decrease in the surprisal per verse compared to the BPE and Morfessor segmentations.

\subsection{Surprisal difference and morphological complexity}\label{sec:performance_diff}

In order to look into the effect of morphological complexity on the relative strength of a given segmentation method, we conducted correlation studies with the difference between the surprisal per verse for pairs of segmentation methods (the $\Delta$ values as defined in \S \ref{sec:metrics}).
We considered only the measures of morphological complexity that were continuous variables (i.e., number of types, TTR, Moving-Average TTR, and Mean Length of Word).

As shown in Table \ref{table6:difference-correlation}, all of the corpus-based statistics were highly correlated to the $\Delta$ values.
The correlations range from moderate to high using Spearman's $\rho$ ($0.50 < \rho < 0.95$).
Even though the strength of correlations varied slightly, number of types, TTR, MATTR, and MLW all showed a similar correlation with the difference statistics. 
They all had a positive correlation with $\Delta_{\,\text{BPE, char}}$. This indicates that the more morphologically complex a language is, the better it is modeled with character segmentation compared to BPE segmentation. 
Similarly, there were positive correlations between the morphological measures and $\Delta_{\,\text{ Morfessor, char}}$, suggesting that  character segmentation works better than Morfessor in modeling morphologically complex languages.
$\Delta_{\,\text{BPE, Morfessor}}$ also had positive correlations with complexity measures. This means that languages with higher morphological complexity tend to record lower surprisal values with Morfessor segmentation than BPE. 
While BPE and Morfessor models outperformed character models on average as shown in \S \ref{sec:Morfessor_segmentation}, the positive correlations with $\Delta_{\,\text{ Morfessor, char}}$ and $\Delta_{\,\text{BPE, char}}$ suggest that character segmentation outperformed BPE and Morfessor segmentation for languages with very rich morphology.

\begin{table}[t]
\centering
 \begin{tabularx}{\columnwidth}{XXS}
\toprule
Difference & Measure & {Spearman's $\rho$}\\
\midrule
\multirow{4}{*}{$\Delta_{\,\text{BPE, char}}$} & Types & 0.95\threeS \\
& TTR & 0.92\threeS\\
& MATTR & 0.77\threeS\\
& MLW & 0.74\threeS\\
\midrule
\multirow{4}{*}{$\Delta_{\,\text{Morfessor, char}}$} & Types & 0.71\threeS\\
& TTR &0.66\threeS\\
& MATTR & 0.50\threeS \\
& MLW & 0.53\threeS \\
\midrule
\multirow{4}{*}{$\Delta_{\,\text{BPE, Morfessor}}$} & Types & 0.86\threeS \\
& TTR & 0.86\threeS\\  
& MATTR & 0.80\threeS\\
& MLW & 0.75\threeS \\
\bottomrule
 \end{tabularx}
\caption{Correlation between surprisal differences and morphological complexity measures for character, BPE, and Morfessor models. All $p$-values $<10^{-11}$.}
\label{table6:difference-correlation} 
\end{table}

\begin{figure*}[t]
    \centering
    \includegraphics[width=\textwidth]{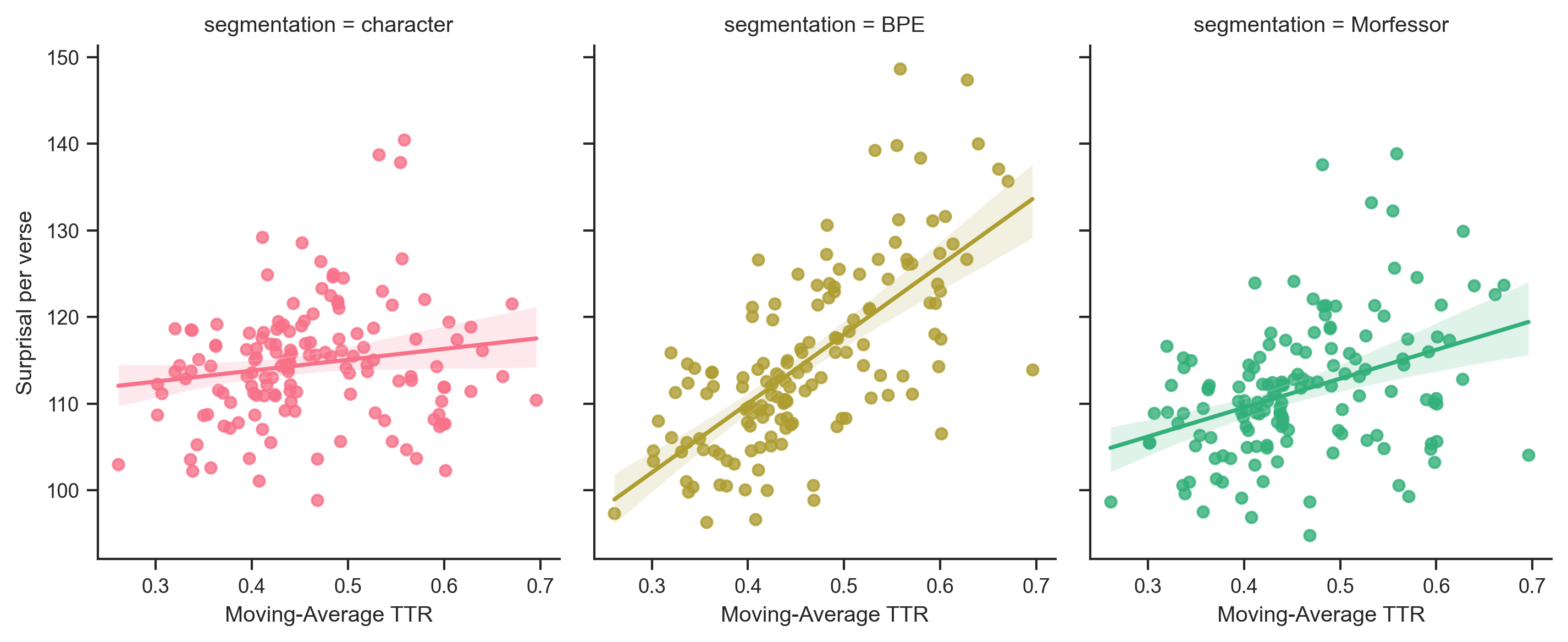}
    \caption{Surprisal per verse plotted against Moving-Average TTR for character, BPE, and Morfessor segmentation methods. Lines indicate the regression estimate with 95\% confidence intervals.}
    \label{figure3:MATTR_surprisal}
\end{figure*}

These results are supported by Figure \ref{figure3:MATTR_surprisal}, where the surprisal per verse for different segmentation models is plotted against MATTR.\footnote{The same trend was captured when we plotted with the other corpus-based measures.} 
For languages with lower MATTR, BPE and Morfessor perform better than character segmentation. However, for languages with higher MATTR, character and Morfessor models outperform BPE.

\section{Discussion}

Our results show that BPE models' surprisal per verse is highly correlated with a language's morphology, represented by several WALS features and corpus-based measures.
Morfessor shows weaker correlations with such measures and records better performance for most of the languages. FST-based models outperform others when available. 
In this section, we discuss the implications of these findings in the context of previous work and future research.

\subsection{Morphology and surprisal}

In accordance with the prior work discussed in \S \ref{sec:lm-review}, we found differences in modeling difficulty between languages. The correlation studies in \S \ref{sec:results} provide evidence that morphology is a substantial contributing factor to these differences. 
Six WALS \citep{wals} morphology features showed association with the surprisal per verse of BPE language models.
Corpus-based statistics like number of types and MATTR showed strong correlations with BPE surprisal, supporting the relationship between modeling difficulty and morphological complexity. 

Our conclusion that a language's morphology impacts language modeling difficulty agrees with \citet{cotterell-etal-2018-languages} and \citet{gerz-etal-2018-relation}, but is at odds with \citet{mielke-etal-2019-kind}.
We included languages known for their rich morphology, such as Western Canadian Inuktitut ({ikt}) and Central Alaskan Yup'ik ({esu}), which may have increased the variation in morphological complexity in the corpus.
We also augmented the WALS data by consulting reference grammars, so we were able to consider 11 more morphological WALS features than \citet{mielke-etal-2019-kind}.
We found that the morphological feature \citet{mielke-etal-2019-kind} considered, 26A ``Prefixing vs. Suffixing in Inflectional Morphology,'' indeed showed no correlation with BPE surprisal.
However, our results show that there are aspects of morphology that affect surprisal that were not considered before.

Previous work, such as \citet{gerz-etal-2018-relation}, focused only on aspects of morphology that they believed \emph{a priori} would predict language model performance. 
In contrast, our study tested all of the morphological features listed in WALS and also tested each of them individually. We found that two of the four features in \citet{gerz-etal-2018-relation}, 20A ``Fusion of Selected Inflectional Formatives'' and 22A ``Inflectional Synthesis of the Verb,'' showed no association with language model performance. Additionally, we found several features that affected language modeling performance, specifically locus of marking and syncretism, which were not mentioned in the literature. These results show that the features tied to morphological complexity in previous work are not necessarily the same features that affect language modeling.

In addition to differences in results, our interpretation of corpus-based statistics like TTR also diverges from previous work.
While \citet{mielke-etal-2019-kind} reported high correlations between language model performance and such statistics, they considered them only as simple statistics of the data. In fact, our results replicate \citet{mielke-etal-2019-kind} in that the number of types was the most predictive of BPE language model surprisal among all the variables considered. 
However, we argue that corpus-based statistics can be used as an approximate measure of morphological complexity based on previous studies.
These corpus-based measures of morphology are reported to capture the overall ranking of morphological complexity \citep{kettunenCanTypeTokenRatio2014a, bentzComparisonMorphologicalComplexity2016c} and can be interpreted in relation to morphological typology \citep{gerz-etal-2018-relation}. We also believe our results indicate that TTR and the WALS features capture similar information.
For example, the positive correlation of $\Delta_{\,\text{BPE, Morfessor}}$ for corpus-based measures corresponds to the smaller effect sizes of WALS features found for Morfessor compared to BPE. This indicates a lesser effect of rich morphology on Morfessor models compared to BPE.

\subsection{Segmentation methods}

While the primary goal of this work is to analyze the relation of a language's morphology to language modeling performance, we found this to be entangled with the level and method of segmentation. Our results show that there is significant variation in the effectiveness of segmentation methods cross-linguistically, and suggest challenges to the status quo methods of subword segmentation in particular. While the subword segmentation methods we used generally outperformed character-level segmentation, 
the higher the TTR, the smaller the difference in surprisal for both BPE and Morfessor, suggesting that these methods are less effective at segmenting languages with highly complex morphology. Of pre-existing methods, we found Morfessor to have the lowest surprisal per verse for most of the languages considered. 
Morfessor's weaker correlations with WALS features and other measures like TTR suggest that its better performance may be due to a better ability to model languages with a wider range of morphological attributes.
This is in line with \citet{Bostrom2020BytePE}, which showed that Unigram LM \cite{kudo-2018-subword}, a segmentation algorithm similar to Morfessor, often outperforms BPE and produces more morph-like segmentation in the context of language model pretraining in English and Japanese.

However, Morfessor was significantly outperformed by character segmentation for a small subset of languages.\footnote{amh, arz, ayr, cmn, esu, heb, ike, ikt, kal, quh, tel, xho. BPE outperformed Morfessor for cmn and heb.}
Many of these languages have been classified as polysynthetic, suggesting that perhaps Morfessor is ill-suited for such languages \cite[see][for discussions on challenges polysynthetic languages pose for NLP tasks]{klavans-2018-computational, tyers-mishchenkova-2020-dependency, mager-etal-2018-lost}.

Additionally, for a typologically diverse subset of languages for which we could obtain FST morphological segmenters, we considered novel segmentation methods: FST$+$BPE and FST$+$Morfessor. We found this simple extension of BPE and Morfessor with morphological information achieved the lowest surprisal per verse in all available languages. 
The overall success of combining statistical segmentations with FSTs further confirms the impact of morphology on language modeling and yields significant promise for the use of segmentation based on linguistic morphological information.

\section{Conclusion}
A language's morphology is strongly associated with language modeling surprisal for BPE-segmented language models. BPE model surprisal is associated with 6 out of the 12 studied WALS morphology features, indicating that there are aspects of some languages' morphology that BPE does not help mitigate.
Strong correlations with corpus-based measures of morphology such as TTR further suggest that the more types available in a language (often by means of rich morphology), the harder it is to model based on BPE units.
Morfessor, which was designed with morpheme induction in mind, performs better for most languages and shows less association with morphological features. When available, the linguistically-informed method of FST-augmented BPE or Morfessor segmentation performs best, indicating a further promise for using linguistic knowledge to combat the effects of morphology on language model surprisal. 

These conclusions were only possible through manual augmentation of typological databases and expansion of studied languages. 
Future efforts could adopt our approach for other areas of language. 
Using linguistically-informed resources across many languages is an avenue for improving neural models in NLP in both design and analysis.

\section*{Acknowledgments}
This paper builds on our prior work for the 2019 Sixth Frederick Jelinek Memorial Summer Workshop on Speech and Language Technology (JSALT 2019) \cite{schwartz2020neural}. We thank the organizers of the workshop and the members of our workshop team on Neural Polysynthetic Language Modeling for inspiring us to pursue this research direction. Our special thanks to Rebecca Knowles, Christo Kirov, Lori Levin, Chi-kiu (Jackie) Lo, and TACL reviewers and editors for their feedback on our manuscript.
We thank Ata Tuncer for his assistance with Turkish segmentation.
This work utilizes resources supported by the National Science Foundation's Major Research Instrumentation program, grant \#1725729, as well as the University of Illinois at Urbana-Champaign.

\bibliography{main-2551-Park}
\bibliographystyle{acl_natbib}

\appendix
\section{Data} \label{Appendix:data}

We began with the data used in \citet{mielke-etal-2019-kind}.
This was originally a subset of a Bible corpus \citep{mayer-cysouw-2014-creating}, which is no longer publically available.
We excluded constructed languages ({epo, tlh}) from the data, keeping a total of 104 verse-aligned Bibles in 60 languages\footnote{{afr, aln, arb, arz, ayr, bba, ben, bqc, bul, cac, 
cak, ceb, ces, cmn, cnh, cym, dan, deu, ell, eng, 
fin, fra, guj, gur, hat, hrv, hun, ind, ita, 
kek, kjb, lat, lit, mah, mam, mri, mya, nld, nor, 
plt, poh, por, qub, quh, quy, quz, ron, rus, som, 
tbz, tel, tgl, 
tpi, tpm, ukr, vie, wal, wbm, 
xho, zom}} in 12 language families.
To increase the number of the languages and language families represented, we added 41 Bibles in 32 languages to the data. 
13 Bible translations in 13 languages\footnote{{als, amh, dje, heb, isl, jpn, kor, pck, slk, slv, 
spa, swe, tha}} were sourced from \citet{bible-Christo-2014}.
In addition, we included 28 Bible translations in 21 languages scraped from various online sources.
Two of the Bibles scraped were in Spanish ({spa}) and Telugu ({tel}), languages which were already included in the Bible corpora \citep{mayer-cysouw-2014-creating, bible-Christo-2014}.
These translations were included because the new Spanish Bible was a parallel source of the Paraguayan Guaraní ({gug}) translation, and the Telugu Bible obtained from \citet{mielke-etal-2019-kind} was originally mislabeled as Tecpatlán Totonac ({tcw}).
The Central Alaskan Yup'ik ({esu}) Bible was from \url{https://bibles.org}.
26 Bibles in 19 languages\footnote{{crk, gug, gui, hin, ike, ikt, kan, mal, mar, nch, nep, nhe, pes, pol, sna, spa, tel, tob, tur}} were from \url{http://bible.com}.
The Greenlandic ({kal}) Bible was obtained from \url{http://old.bibelselskabet.dk}.

\end{document}